\documentclass[runningheads]{llncs}
\usepackage[T1]{fontenc}
\usepackage{graphicx}
\usepackage{float}
\usepackage{algorithm,algorithmic,amsmath}
\usepackage{multirow}
\usepackage{array}

\usepackage{booktabs}
\usepackage{boldline}
\usepackage{adjustbox}
\usepackage{makecell}

\newcolumntype{C}[1]{>{\centering\arraybackslash}p{#1}}
\usepackage[hidelinks,bookmarksopen=true,bookmarksnumbered=true]{hyperref}
\begin{document}
\title{Distill-then-Replace: Efficient Task-Specific Hybrid Attention Model Construction}
\titlerunning{Efficient Task-Specific Hybrid Attention Model Construction}
\author{Xiaojie Xia\inst{1}\orcidID{0000-0002-6486-7557} \and
Huigang Zhang\inst{1}\orcidID{0000-0002-7567-7177} \and
Chaoliang Zhong\inst{1}\orcidID{0000-0003-1697-6932} \and
Jun Sun\inst{1}\orcidID{0000-0002-0967-4859} \and
Yusuke Oishi\inst{2}\orcidID{0000-0003-4264-8932}
}
%
% \authorrunning{Xiaojie Xia et al.}
%\institute{Princeton University, Princeton NJ 08544, USA \\

\institute{Fujitsu Research \& Development Center CO., LTD, China\\
\email{\{xiaxiaojie,zhanghuigang,clzhong,sunjun\}@fujitsu.com} \and
Fujitsu Research, FUJITSU LTD, Japan\\
\email{\{oishi.y@fujitsu.com}\}
}

\maketitle              % typeset the header of the contribution
\begin{abstract}
Transformer architectures deliver state-of-the-art accuracy via dense full-attention, but their quadratic time and memory complexity with respect to sequence length limits practical deployment.
Linear attention mechanisms offer linear or near-linear scaling yet often incur performance degradation.
Hybrid models that integrate full and linear attention layers promise a balance between efficiency and expressiveness, but face two major challenges: training such hybrid models from scratch is computationally expensive, and manually designing the optimal placement of attention types is highly nontrivial.
We propose \textbf{DtR} (\textbf{D}istill-\textbf{t}hen-\textbf{R}eplace), which first transfers weights from the pretrained full-attention modules to its linear attention counterparts through blockwise local distillation, and then applies a greedy layer replacement strategy that iteratively substitutes full attention blocks with linear ones while monitoring validation performance on the target task. 
DtR yields a task-specific hybrid model in a single efficient pass, without costly re-training or neural architecture search, and can be applied to any pretrained full-attention backbone for diverse downstream tasks.

\end{abstract}
\section{Introduction}
%Transformer architectures \cite{vaswani2017attention} have achieved state-of-the-art performance across numerous domains, primarily driven by their dense, softmax-based self-attention mechanism. However, this core component suffers from quadratic time and memory complexity with respect to sequence length, posing a significant constraint for deployment, especially on some long-context tasks such as document understanding and extended dialogue \cite{liu2025survey,bulatov2024beyond,tang2024survey}. 

Transformer architectures\cite{vaswani2017attention} achieve state-of-the-art performance through dense, softmax-based self-attention, yet suffer from quadratic time and memory complexity with respect to sequence length. This poses significant constraints for long-context tasks such as document understanding and extended dialogue\cite{liu2025survey}.

%To address this challenge, recent research efforts have focused on developing efficient alternatives, particularly attention mechanisms that operate with linear or near-linear complexity \cite{gu2021efficiently,katharopoulos2020transformers,peng2023rwkv}. Models such like RetNet \cite{sun2023retentive}, Mamba \cite{gu2024mamba} and Gated DeltaNet \cite{yanggated} compress the linearly growing Key-Value Cache(KV-Cache) into a static hidden state, thereby significantly reducing both memory usage and computational overhead \cite{wang2025systematic}. While these methods offer substantial gains in efficiency and scalability, they often incur a performance penalty, particularly in tasks requiring precise recall or nuanced contextual understanding \cite{waleffe2024empirical,qu2024survey,arora2024simple}, due to the compression of historical information.

Recent works have developed efficient alternatives with linear or near-linear complexity \cite{gu2021efficiently,katharopoulos2020transformers,peng2023rwkv}. Models such as RetNet \cite{sun2023retentive}, Mamba \cite{gu2024mamba}, and Gated DeltaNet \cite{yanggated} compress the KV Cache into a static hidden state, substantially reducing memory and computation. However, these methods often incur performance penalties on tasks requiring precise recall or nuanced contextual understanding \cite{waleffe2024empirical}, due to historical information compression.

%Hybrid architectures, which strategically interleave standard full-attention blocks with linear-attention blocks, present a promising direction to balance computational efficiency and model expressivity \cite{ren2024samba,glorioso2024zamba,lenz2025jamba}. In principle, such hybrids can preserve the high accuracy of transformers on critical tokens or segments while achieving linear scaling for the majority of the sequence. Despite this promise, two major practical barriers impede the widespread adoption of hybrid models. First, training such a hybrid model from scratch requires computational resources and data on par with pretraining a full-scale transformer from scratch, making it prohibitively costly in most scenarios. Second, determining the optimal placement of full-attention and linear-attention layers for the model architecture is a complex and non-trivial engineering challenge, for which no generalizable or systematic solution currently exists.

Hybrid architectures that interleave full-attention and linear-attention blocks balance efficiency and expressivity \cite{ren2024samba,glorioso2024zamba,lenz2025jamba}, preserving transformer accuracy on critical tokens while achieving linear scaling elsewhere. Yet two barriers impede adoption: (i) training from scratch requires resources comparable to full-scale pretraining, making it prohibitively costly; and (ii) optimal placement of attention types is a complex engineering challenge with no systematic solution.

% In this work, we attempt to develop a method to efficient, task-specific creation of high-performance hybrid models from existing pretrained transformers. Our proposed approach transferring linear layer weights from a pretrained full-attention transformer via block-wise local distillation, avoiding expensive re-pretraining, and applying a greedy layer-replacement algorithm that incrementally substitutes full-attention blocks with their linear counterparts while monitoring task-specific validation performance. The resulting task-specific hybrid model is produced in a single low-cost pass, requires no costly architectural search, and can be instantiated from any pretrained full-attention backbone for various downstream tasks.

We address both issues by proposing efficient, task-specific hybrid model creation from existing pretrained transformers. 
Our approach, \textbf{DtR}  (\textbf{D}istill-\textbf{t}hen-\textbf{R}eplace), transfers weights to linear counterparts via blockwise local distillation, avoiding expensive re-pretraining, then applies a greedy layer-replacement algorithm that incrementally substitutes full-attention blocks with linear ones while monitoring task-specific validation performance. DtR produces the resulting hybrid model in a single low-cost pass, requires no costly architectural search, and can be instantiated from any pretrained full-attention backbone for diverse downstream tasks.

\section{Related works}
\subsection{Linear Complexity Models}

While Transformers \cite{vaswani2017attention} have become the de facto standard for LLMs, their global attention suffers from quadratic complexity $O(N^2)$. Early works proposed kernel-based linear attention mechanisms, such as Linear Transformer \cite{katharopoulos2020transformers} and Performer \cite{choromanski2020rethinking}, which approximate the softmax attention map. More recently, architectures unify Transformers' parallel training with RNNs' efficient inference. RWKV \cite{peng2023rwkv}  reformulates linear attention into a recurrent form for linear inference scaling. RetNet \cite{sun2023retentive} introduces multi-scale exponential decay retention, achieving training parallelism, low-cost inference, and strong performance. Structured state-space models such as Mamba \cite{gu2024mamba,dao2024transformers} incorporate data-dependent state transitions for high expressivity and subquadratic scaling. Gated Linear Attention \cite{yang2024gated}and Gated DeltaNet \cite{yanggated} integrate gating mechanisms into linear recurrent units, enhancing stability and capacity. These models demonstrate that efficient attention alternatives can achieve competitive performance.

\subsection{Hybrid Architectures}
%Pure linear architectures sometimes struggle with tasks requiring high-fidelity associative recall or in-context retrieval compared to standard attention. To bridge this gap, hybrid models combine the strengths of both paradigms. H3 \cite{fu2022hungry} was among the first to effectively bridge the gap between SSMs and transformers on language tasks. More recently, Jamba \cite{lenz2025jamba}, Samba \cite{ren2024samba}, and Zamba \cite{glorioso2024zamba} have pushed hybrid scaling to new heights by weaving Mamba layers with transformer blocks in distinct and complementary ways, all attaining state-of-the-art results. Similarly, Google's Griffin\cite{de2024griffin} mixes gated linear recurrences with local attention layers, demonstrating that a small budget of global attention is sufficient to recover the capabilities of full-attention transformers. Jet-Nemotron \cite{gu2025jet} further pushes the efficiency frontier by employing a post-training neural architecture search to automatically discover the optimal placement of attention and linear recurrence blocks, yielding models that match transformer quality while reducing inference latency. These works suggest that hybridizing linear layers with standard attention is a promising direction for next-generation LLMs.

Pure linear architectures sometimes struggle with high-fidelity associative recall and in-context retrieval. Hybrid models bridge this gap by combining both paradigms. H3 \cite{fu2022hungry}  first effectively connected SSMs with transformers on language tasks. More recently, Jamba \cite{lenz2025jamba}, Samba \cite{ren2024samba}, and Zamba \cite{glorioso2024zamba} wove Mamba layers with transformer blocks in complementary ways, attaining state-of-the-art results. Griffin \cite{de2024griffin} mixes gated linear recurrences with local attention, showing that limited global attention suffices to recover full-attention capabilities. Jet-Nemotron \cite{gu2025jet} employs post-training neural architecture search to discover optimal attention placement automatically. These works establish hybridization as a promising direction. However, these hybrid models typically require substantial training 
resources and extended training time.

\subsection{LLM Distillation}
% Knowledge Distillation (KD) is widely used to compress large teacher models into smaller student models. While traditional KD \cite{hinton2015distilling} focuses on matching logits or hidden states, distilling generative LLMs presents unique challenges due to the sequential nature of text generation. MiniLLM \cite{gu2023minillm} identifies that minimizing the forward Kullback-Leibler (KL) divergence leads to the student overestimating low-probability regions of the teacher, and proposes using reverse KL divergence instead. \cite{agarwal2024policy} explores generalized knowledge distillation, investigating the impact of data selection (e.g., on-policy vs. off-policy) on distillation efficiency. Furthermore, recent research \cite{wang2024mamba,yang2025zebra,goldstein2025radlads} has explored cross-architecture distillation, where a transformer teacher distills knowledge into a linear or hybrid student (e.g., Mamba-2), ensuring the efficient student inherits the strong reasoning capabilities of the standard Transformer.

Knowledge Distillation (KD) \cite{hinton2015distilling} compresses large teachers into smaller students. For generative LLMs, MiniLLM \cite{gu2023minillm}identifies that forward KL divergence causes students to overestimate low-probability regions, proposing reverse KL instead. \cite{agarwal2024policy} explores generalized distillation through data selection strategies. Recent works \cite{yang2025zebra,goldstein2025radlads} investigate cross-architecture distillation, where transformer teachers distill into linear or hybrid students (e.g., Mamba-2), ensuring efficient students inherit strong reasoning capabilities. Yet cross-architecture distillation remains data-intensive and 
computationally demanding.

\vspace{-5pt}
\section{Methodology}
Our proposed DtR (Distill-then-Replace) introduces a streamlined, cost-effective pipeline for converting any pretrained, standard transformer into a task-specific hybrid model with an optimized mixture of full- and linear-attention layers. It consists of two core components: weight transferring via blockwise local distillation and an automated greedy layer-replacement algorithm.
\vspace{-5pt}
\subsection{Blockwise Local Distillation to Linear Attention Counterparts}
In this stage, our objective is to derive a linear attention mechanism that can effectively approximate the behavior of each softmax-based full-attention block in the original transformer architecture. Inspired by the methodology introduced in \cite{bercovich2024puzzle}, we train each linear counterpart independently, such that it accurately reproduces the output of its corresponding full-attention module when linear modules are provided with the same hidden state input as the parent block. 

\begin{figure}[!htbp]
\centering
\includegraphics[width=\textwidth]{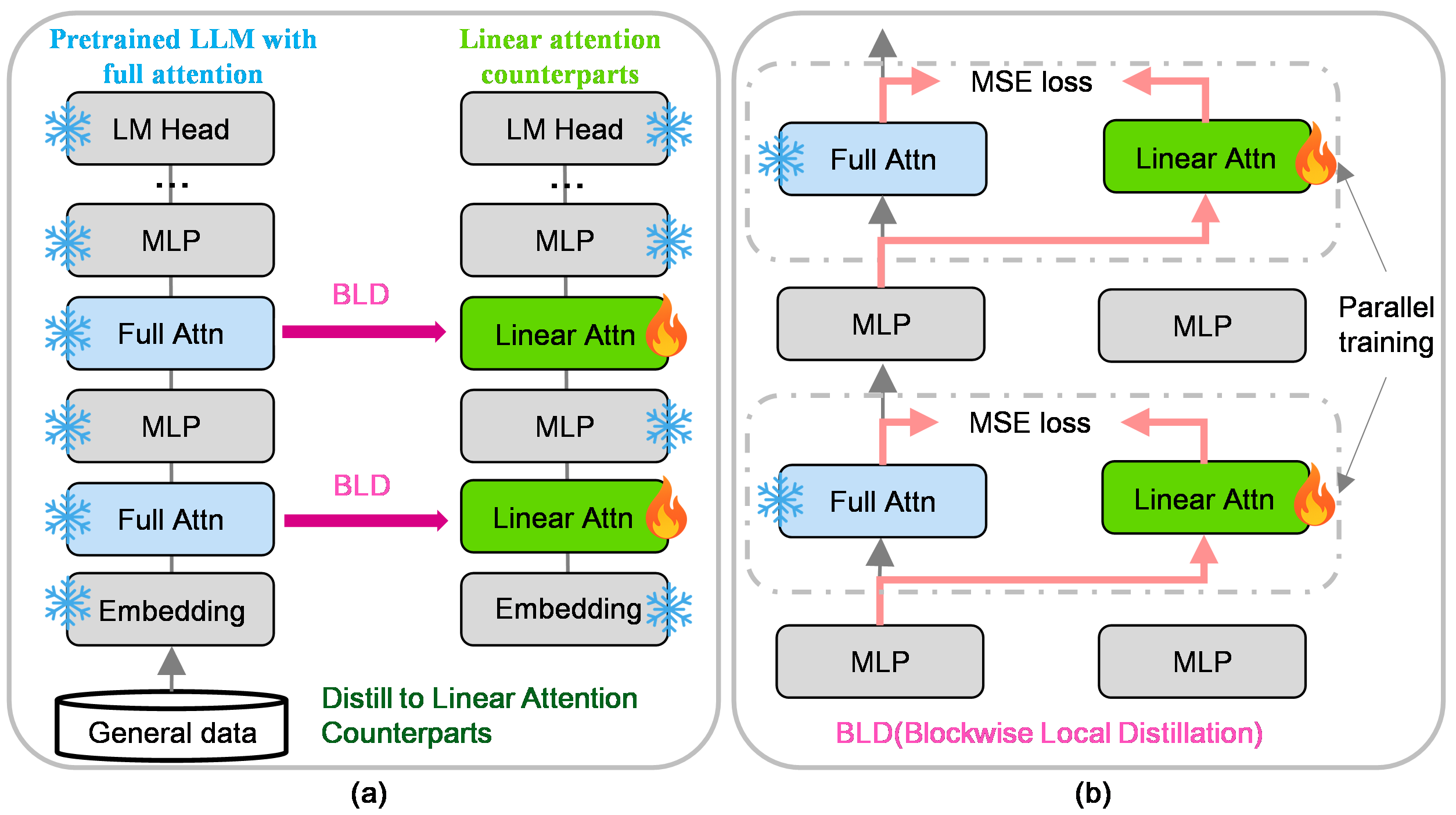}
\caption{Linear attention weight by blockwise local distillation. (a) Overall distillation framework from full attention to linear attention. (b) BLD (blockwise local distillation), which are trained in parallel and independently.} \label{fig1}
\end{figure}

As illustrated in Fig.~\ref{fig1}, the block-wise local distillation objective $\mathcal{L}$ is concisely expressed as

\[
\mathcal{L} = \mathrm{MSE}\!\left(O_{\text{full}},\, O_{\text{linear}}\right)
\]
where $O_{\text{full}}$ and $O_{\text{linear}}$ represent the outputs of the original full-attention block and the linear counterpart block, respectively.

This decoupled distillation ensures that each linear attention module depends solely on the behavior of its associated full-attention block, without requiring back-propagation through the entire network. Consequently, all linear attention modules can be trained in parallel, significantly reducing computational overhead and accelerating convergence compared to end-to-end joint pretraining. This design not only enhances training efficiency but also preserves the functional fidelity of the original attention mechanism, facilitating seamless integration into existing architectures.

\vspace{-10pt}

\subsection{Task-Specific Hybrid Architecture Construction}
Upon obtaining both the original full-attention transformer model and its distilled linear counterparts, we now seek an optimal hybrid architecture that balances performance and efficiency for a specific downstream task. Rather than resorting to costly neural architecture search  or end-to-end fine-tuning of hybrid configurations, we propose a greedy, validation-driven layer replacement strategy that constructs a task-specific hybrid model in a single, low-overhead pass.

%Formally, let \(\mathcal{M}_\text{full}\) denote the pretrained model composed of \(L\) full-attention blocks, and let \(\mathcal{T}\) represent the downstream task with a validation set. We first evaluate the baseline performance of \(\mathcal{M}_\text{full}\) on \(\mathcal{T}\) using a task-specific metric (e.g., accuracy or F1-score), denoted as \(\mathcal{P}(\mathcal{M}_\text{full}; \mathcal{T})\). A minimum acceptable performance threshold \(\mathcal{P}_\text{min}\) is then established. Since linear modules eliminate the need for the corresponding full-attention module’s KV-Cache, replacing more full-attention layers to linear counterparts can substantially improve the overall throughput of the model. Our objective is to derive a hybrid model \(\mathcal{M}_\text{hybrid}\) that replaces as many full-attention blocks as possible with their linear counterparts, while still satisfying the performance constraint:
%\[
%\mathcal{P}(\mathcal{M}_\text{hybrid}; \mathcal{T}) \geq \mathcal{P}_\text{min}
%\]
%We halt at the moment this condition is violated; the hybrid model produced up to that point is taken as the task-optimal model $\mathcal{M}_\text{opt}$. 
%We also record $\mathcal{M}_\text{best}$, the hybrid model achieving the highest validation performance.
%Searching for an architecture configuration with a fixed number of replaced layers is also viable.
%The complete method is presented in Algorithm \ref{alg1}. The final output is a task-specific hybrid model that retains high performance where it matters most and achieves significant computational savings elsewhere.

Given a pretrained full-attention model $\mathcal{M}_{\mathrm{full}}$ with $L$ blocks and a downstream task $\mathcal{T}$, we construct an efficient hybrid model through greedy layer replacement.
\vspace{-5pt}
\paragraph{Setup.} We evaluate the baseline performance $\mathcal{P}(\mathcal{M}_{\mathrm{full}}; \mathcal{T})$ on a task validation set and establish a minimum tolerated acceptable threshold $\mathcal{P}_{\min}$. The goal is to replace as many full-attention layers with their distilled linear counterparts as possible, while maintaining $\mathcal{P}(\mathcal{M}_{\mathrm{hybrid}}; \mathcal{T}) \geq \mathcal{P}_{\min}$.

\vspace{-5pt}
\paragraph{Greedy Layer Replacement.} Starting from $\mathcal{M}_{\mathrm{full}}$, we iteratively substitute one full-attention block at a time with its linear counterpart and evaluate validation performance. In each iteration, we select the layer whose replacement yields the highest validation score $\mathcal{P}^*$. If $\mathcal{P}^* \geq \mathcal{P}_{\min}$, we commit the replacement and continue; otherwise, we stop. The procedure requires only forward passes---no back-propagation or retraining.

\vspace{-5pt}
\paragraph{Output models.} We record two models: (i) $\mathcal{M}_{\mathrm{best}}$, the hybrid achieving the highest validation performance throughout the search, and (ii) $\mathcal{M}_{\mathrm{opt}}$, the hybrid with the maximum number of linear layers while satisfying the performance constraint. %The entire search completes in a single efficient pass without costly architecture search or retraining.

\vspace{5pt}
The full procedure is presented in Algorithm~\ref{alg1}. Our greedy layer replacement algorithm is motivated by the observation that full and linear attention mechanisms exhibit task-dependent representational strengths. Consequently, where linear modules are placed significantly affects performance: some layers can be replaced without loss—and sometimes with gain—while others are critical to preserve. By iteratively evaluating validation performance after each candidate replacement, our method adapts to each task’s unique knowledge distribution, identifying an effective hybrid configuration without exhaustive search. 
The procedure requires only a forward pass per candidate layer, making it highly efficient.
Crucially, it automatically retains full-attention layers essential for expressivity and replaces less sensitive ones with linear approximations, achieving substantial speedups while maintaining or even improving task performance.

Importantly, our framework is backbone-agnostic and task-general: it can be applied to any pretrained full-attention transformer (e.g., GPT, LLaMA, Qwen) and adapted to various downstream tasks without modifications to the core algorithm. This enables rapid deployment of efficient, high-performing models at a fraction of the cost of training hybrid architectures from scratch or conducting exhaustive architecture search.

\begin{algorithm}[t]
\small
\caption{DtR (Distill-then-Replace): Efficient Task-Specific Hybrid Attention Model Construction}
\label{alg1}
\begin{algorithmic}[1]
\REQUIRE Full-attention model $\mathcal{M}_{full}$ with $L$ blocks; Training data $\mathcal{D}$; 
         Downstream task training dataset $\mathcal{T}_{test}$; Downstream task validation set $\mathcal{T}_{val}$; Metric $\mathcal{P}(\cdot; \mathcal{T})$; 
         Minimum performance threshold $\mathcal{P}_{min}$
\ENSURE $\mathcal{M}_{best}$: highest-performance hybrid model; $\mathcal{P}_{best}$: its metric; 
        $\mathcal{M}_{opt}$: maximum-replacement hybrid with $\mathcal{P} \geq \mathcal{P}_{min}$; $\mathcal{P}_{opt}$: its metric
\STATE \textbf{Phase 1:} BLD(Blockwise Local Distillation) to Linear Attention Counterparts
\FOR{$\ell = 1$ to $L$ \textbf{in parallel}}
    \STATE $\mathcal{M}_{linear}^{(\ell)} \leftarrow \text{BLD}(\mathcal{M}_{full}^{(\ell)}; \mathcal{D})$
\ENDFOR
\STATE \textbf{Phase 2:} Greedy Layer Replacement to Hybrid Attention Model
\STATE $\mathcal{M} \leftarrow \mathcal{M}_{full}$, $\mathcal{P}_{best} \leftarrow \mathcal{P}(\mathcal{M}; 
\mathcal{T}_{val})$, $\mathcal{M}_{best} \leftarrow \mathcal{M}$, $\mathcal{M}_{opt} \leftarrow \mathcal{M}$, $\mathcal{P}_{opt} \leftarrow \mathcal{P}(\mathcal{M}; \mathcal{T}_{val})$
\REPEAT
    \STATE $\mathcal{M}^*, \mathcal{P}^*, \ell^* \leftarrow \emptyset, -\infty, -1$
    \FOR{each full-attention layer index $\ell$ in $\mathcal{M}_{full}$}
        \STATE $\mathcal{M}_\ell \leftarrow \text{Replace-Layer-to-Linear}(\mathcal{M}, \ell, \mathcal{M}_{linear}^{(\ell)})$
        \STATE $\mathcal{P}_\ell \leftarrow \mathcal{P}(\mathcal{M}_\ell; \mathcal{T}_{val})$
        \STATE $(\mathcal{M}^*, \mathcal{P}^*, \ell^*) \leftarrow (\mathcal{M}_\ell, \mathcal{P}_\ell, \ell)$ \textbf{if} $\mathcal{P}_\ell > \mathcal{P}^*$
    \ENDFOR
    \STATE $\mathcal{M}_{best} \leftarrow M^*$, $\mathcal{P}_{best} \leftarrow \mathcal{P}^*$ \textbf{if} $\mathcal{P}^* \geq \mathcal{P}_{best}$
    \STATE $\mathcal{M} \leftarrow \mathcal{M}^*$, $M_{opt} \leftarrow \mathcal{M}$, $\mathcal{P}_{opt} \leftarrow \mathcal{P}^*$ 
           \textbf{if} $\mathcal{P}^* \geq \mathcal{P}_{min}$ \textbf{else break}
\UNTIL{$\mathcal{P}^* < \mathcal{P}_{min}$}
\STATE $\mathcal{P}_{best}^{test} \leftarrow \mathcal{P}(\mathcal{M}_{best}; \mathcal{T}_{test})$, $\mathcal{P}_{opt}^{test} \leftarrow \mathcal{P}(\mathcal{M}_{opt}; \mathcal{T}_{test})$
\STATE \textbf{return} $\mathcal{M}_{best}$, $\mathcal{P}_{best}^{test}$, $\mathcal{M}_{opt}$, $\mathcal{P}_{opt}^{test}$
\end{algorithmic}
\end{algorithm}
\vspace{-15pt}

\section{Experiments}
\vspace{-10pt}
Having introduced our proposed DtR (Distill-then-Replace), we now present a series of experiments designed to build a comprehensive case for its effectiveness. 

\subsection{Experiment Setup}
\subsubsection{Datasets, Models and Evaluation Details.}

We evaluate our method across a diverse suite of benchmarks spanning commonsense knowledge, reasoning, code generation, information extraction, long-context summarization and specific domain tasks.
For commonsense and reasoning tasks, we consider ArcC \cite{clark2018think}, ArcE \cite{clark2018think}, BoolQ(BQ) \cite{clark2019boolq}, CommonsenseQA(CQ) \cite{talmor2019commonsenseqa}, HellaSwag(HS) \cite{zellers2019hellaswag}, OpenbookQA (OBA) \cite{mihaylov2018can},  PIQA \cite{bisk2020piqa} and Winogrande(WG)  \cite{sakaguchi2021winogrande}. 
For mathematical reasoning, we include GSM8K \cite{cobbe2021training} and MathQA(MQ) \cite{amini2019mathqa}.
For code generation, we evaluate on HumanEval(HE) \cite{chen2021evaluating} and MBPP \cite{austin2021program}. 
For information extraction, we include SWDE \cite{arora2024simple}. 
For long-context evaluation, we include SAMSum, a dialogue summarization subtask, and TREC, a question classification subtask, both from LongBench \cite{bai2024longbench}.
For specific domain tasks, we include two MMLU \cite{hendrycks2020measuring} subtasks: econometrics(EC) and marketing(MK), and PubMedQA(PM) \cite{jin2019pubmedqa}.
We adopt the official validation split as the evaluation dataset in our algorithm if available; otherwise, we randomly sample a small subset from the training set.

We conduct experiments on three decoder-only base models: Qwen2.5-1.5B \cite{team2024qwen2}, Llama-3.2-3B-Instruct \cite{grattafiori2024llama}, and Llama-3.1-8B-Instruct \cite{grattafiori2024llama}, with 28, 28, and 32 layers, respectively. We evaluate using LM-Evaluation-Harness \cite{eval-harness} following a zero-shot protocol (except for GSM8K, which adopt 4-shot evaluation), reporting each dataset's official metrics, such as accuracy (acc), normalized accuracy (acc\_norm), pass\@1, F1-score, or Rouge-L, where higher values indicate better performance.

\subsubsection{Method Details.}
Our method consists of two stages. 
In the blockwise local distillation stage, we randomly sample about 100M mini-corpus from a combination of Nemotron-CC \cite{su2025nemotron} and Redstone-QA \cite{chang2024redstone}. We only train the linear modules using a MSE loss while keeping the embeddings, MLPs and LM head frozen.
During layer replacement stage, we first evaluate the full-attention model and record its performance as a baseline, then iteratively replace layers to linear attention on each task’s validation set to obtain the best architecture, and then evaluate on their test split.
The alternative linear modules evaluated are Gated DeltaNet(GDN) \cite{yanggated}, Gated Linear Attention(GLA) \cite{yang2024gated}  and JetBlock(JET) \cite{gu2025jet}, all configured with the same hidden size as corresponding full-attention layers. The entire procedure is executed on a single NVIDIA A800 GPU. 

\subsection{Results}
\subsubsection{DtR-searched Best Hybrid Model Performance.}
We begin by evaluating the performance of the best hybrid model across various linear attention variants, as shown in Table \ref{tab1}, using different pretrained base models(Qwen2.5-1.5B, Llama-3.2-3B-Instruct and Llama-3.1-8B-Instruct).
$\mathcal{P}_{\text{base}}$ and $\mathcal{P}_{\text{best}}$ denote the performance of the base and searched best hybrid models on each test dataset, respectively; $\#R$ is the number of replaced layers to linear attention. The performance in \textbf{bold} indicates that the searched hybrid model matches or exceeds the pretrained base model on the corresponding task.

\begin{table}[!htbp]
\centering
\caption{Performance of base models (\textbf{Qwen2.5-1.5B}, \textbf{Llama-3.2-3B-Instruct} and \textbf{Llama-3.1-8B-Instruct}) and best DtR-searched hybrid models with replaced linear modules (\textbf{GLA}, \textbf{GDN} and \textbf{JET}) across tasks.}
\label{tab1}
\begin{adjustbox}{max width=\textwidth,center}
\begin{tabular}{lccccccccccccccccccccc} %{|l|c|cc|cc|cc|c|cc|cc|cc|c|cc|cc|cc|}
\hline
\multirow{3}{*}{\textbf{Datasets}} & 
\multicolumn{7}{c}{\textbf{Qwen2.5-1.5B}} & 
\multicolumn{7}{c}{\textbf{Llama-3.2-3B-Instruct}} & 
\multicolumn{7}{c}{\textbf{Llama-3.1-8B-Instruct}} \\
\cline{2-22}
& \multirow{2}{*}{$\mathcal{P}_{\text{base}}$} & 
\multicolumn{2}{c}{\textbf{GLA}} & 
\multicolumn{2}{c}{\textbf{GDN}} & 
\multicolumn{2}{c}{\textbf{JET}} & 
%\multicolumn{2}{c|}{\textbf{GLA}} & 
%\multicolumn{2}{c|}{\textbf{GDN}} & 
%\multicolumn{2}{c|}{\textbf{JET}} & 
\multirow{2}{*}{$\mathcal{P}_{\text{base}}$} & 
\multicolumn{2}{c}{\textbf{GLA}} & 
\multicolumn{2}{c}{\textbf{GDN}} & 
\multicolumn{2}{c}{\textbf{JET}} & 
\multirow{2}{*}{$\mathcal{P}_{\text{base}}$} & 
\multicolumn{2}{c}{\textbf{GLA}} & 
\multicolumn{2}{c}{\textbf{GDN}} & 
\multicolumn{2}{c}{\textbf{JET}} \\
\cline{3-4} \cline{5-6} \cline{7-8}
\cline{10-11} \cline{12-13} \cline{14-15}
\cline{17-18} \cline{19-20} \cline{21-22}
& & $\mathcal{P}_{\text{best}}$ & \#R & $\mathcal{P}_{\text{best}}$ & \#R & $\mathcal{P}_{\text{best}}$ & \#R & & $\mathcal{P}_{\text{best}}$ & \#R & $\mathcal{P}_{\text{best}}$ & \#R & $\mathcal{P}_{\text{best}}$ & \#R & & $\mathcal{P}_{\text{best}}$ & \#R & $\mathcal{P}_{\text{best}}$ & \#R & $\mathcal{P}_{\text{best}}$ & \#R \\
\hline
ArcC & 45.1 & \textbf{45.3} & 1 & \textbf{45.7} & 3 & \textbf{45.2} & 2 & 46.4 & 44.7 & 6 & 43.1 & 2 & \textbf{46.5} & 4 & 55.0 & \textbf{55.4} & 2 & \textbf{57.1} & 7 & \textbf{55.5} & 2 \\
ArcE & 72.0 & \textbf{74.0} & 2 & \textbf{74.4} & 4 & \textbf{74.7} & 2 & 67.9 & \textbf{69.8} & 5 & \textbf{70.2} & 8 & \textbf{70.8} & 4 & 79.6 & \textbf{80.7} & 7 & \textbf{81.5} & 9 & \textbf{80.7} & 2 \\
BQ & 73.0 & \textbf{75.4} & 3 & \textbf{76.0} & 3 & \textbf{75.2} & 3 & 78.5 & \textbf{80.2} & 9 & \textbf{80.3} & 11 & \textbf{80.5} & 10 & 84.2 & \textbf{85.1} & 13 & \textbf{84.7} & 12 & \textbf{84.6} & 4 \\
CQ & 75.2 & 75.1 & 1 & 74.8 & 1 & 74.1 & 4 & 67.7 & \textbf{68.4} & 6 & 64.8 & 7 & \textbf{68.2} & 3 & 77.2 & 75.1 & 2 & 75.7 & 2 & \textbf{77.3} & 10 \\
HS & 50.1 & \textbf{50.2} & 2 & \textbf{50.4} & 2 & \textbf{52.9} & 4 & 52.4 & \textbf{52.4} & 3 & \textbf{52.4} & 1 & \textbf{54.8} & 3 & 59.2 & \textbf{59.3} & 2 & \textbf{59.3} & 1 & \textbf{59.3} & 2 \\
OBA & 40.6 & \textbf{40.8} & 2 & 40.0 & 2 & \textbf{40.8} & 8 & 36.0 & \textbf{36.2} & 10 & \textbf{37.0} & 3 & \textbf{37.6} & 2 & 43.4 & \textbf{43.8} & 4 & \textbf{43.4} & 5 & \textbf{44.4} & 8 \\
PIQA & 75.7 & \textbf{75.7} & 3 & 75.3 & 2 & 75.0 & 7 & 75.5 & 73.9 & 6 & 75.0 & 2 & \textbf{75.8} & 8 & 80.0 & \textbf{80.1} & 1 & 79.7 & 2 & \textbf{80.1} & 1 \\
WG & 63.3 & \textbf{63.7} & 3 & \textbf{63.3} & 5 & \textbf{64.2} & 6 & 68.0 & 67.2 & 4 & \textbf{68.0} & 5 & \textbf{68.1} & 4 & 74.0 & \textbf{74.2} & 6 & \textbf{75.1} & 10 & \textbf{74.5} & 2 \\
GSM8K & 61.4 & 61.3 & 3 & 60.9 & 1 & 60.1 & 1 & 63.9 & 63.8 & 1 & 63.6 & 1 & 63.2 & 1 & 76.0 & \textbf{76.3} & 2 & \textbf{77.0} & 2 & 75.1 & 1 \\
MQ & 35.0 & \textbf{36.4} & 3 & \textbf{36.3} & 2 & \textbf{35.8} & 3 & 34.5 & \textbf{35.7} & 4 & \textbf{35.8} & 4 & \textbf{35.3} & 3 & 39.4 & \textbf{41.0} & 4 & \textbf{41.1} & 7 & \textbf{40.8} & 3 \\
HE & 47.6 & \textbf{48.2} & 1 & \textbf{48.2} & 1 & \textbf{47.6} & 1 & 51.2 & \textbf{54.9} & 1 & \textbf{53.1} & 1 & \textbf{54.3} & 1 & 58.5 & \textbf{62.8} & 1 & \textbf{65.2} & 4 & \textbf{64.6} & 1 \\
MBPP & 50.0 & \textbf{52.9} & 2 & \textbf{52.7} & 2 & \textbf{52.7} & 2 & 54.8 & \textbf{58.2} & 2 & \textbf{57.1} & 2 & \textbf{56.6} & 1 & 59.8 & \textbf{64.3} & 2 & \textbf{63.8} & 2 & \textbf{62.9} & 2 \\
SWDE & 86.4 & \textbf{89.3} & 5 & \textbf{89.4} & 3 & \textbf{89.1} & 3 & 85.8 & \textbf{89.0} & 4 & \textbf{87.7} & 5 & \textbf{88.4} & 4 & 91.1 & \textbf{92.3} & 5 & \textbf{92.4} & 8 & \textbf{91.6} & 3 \\
SAMSum & 42.5 & \textbf{44.4} & 4 & \textbf{43.9} & 4 & \textbf{43.8} & 3 & 42.4 & \textbf{42.5} & 2 & 42.3 & 1 & \textbf{42.9} & 2 & 43.9 & \textbf{45.1} & 1 & \textbf{45.1} & 3 & \textbf{44.5} & 1 \\
TREC & 71.0 & \textbf{75.0} & 4 & \textbf{76.0} & 5 & \textbf{76.0} & 5 & 72.0 & \textbf{76.0} & 4 & \textbf{75.0} & 5 & \textbf{76.0} & 5 & 73.5 & \textbf{77.0} & 5 & \textbf{77.5} & 4 & \textbf{78.0} & 6 \\
EC	 & 46.5  & 	45.6  & 1	& 43.9  & 	1 & 	\textbf{46.5} 	 & 2 & 	39.5  & 	\textbf{39.5}  & 	1	 & 36.0 	 & 5	 & 39.3  & 	2	 & 51.8 	 & \textbf{53.5}  & 	1 & 	50.9  & 1 &  \textbf{53.5}  & 1 \\
MK & 	83.3  & 	82.5  & 2	 & \textbf{83.8}  & 	4 & \textbf{83.8}  & 	1	 & 86.8 & 	82.5 & 	2 & 	\textbf{86.8}  & 	2 & 	\textbf{87.2}  & 	1 & 	88.9 & \textbf{90.2}  & 	1	 & \textbf{89.3}  & 7	 & \textbf{89.7} 	 & 4 \\
PM & 63.8 & \textbf{72.2} & 7 & \textbf{72.2} & 9 & \textbf{68.4} & 6 & 69.8 & \textbf{72.2} & 7 & \textbf{72.0} & 6 & \textbf{73.0} & 6 & 75.0 & \textbf{78.0} & 13 & \textbf{77.6} & 12 & \textbf{78.4} & 8 \\
\hline
\end{tabular}
\end{adjustbox}
\end{table}

The results demonstrate strong empirical evidence for the effectiveness 
of task-specific hybrid architectures. Across all configurations 
(3 base models $\times$ 3 linear variants $\times$ 18 tasks = 162 settings), 
\textbf{81.5\% (132/162)} achieve performance comparable to or exceeding 
the full-attention baseline. More importantly, \textit{every task admits 
at least one hybrid configuration that matches or surpasses the base 
model}, confirming that optimal placement is task-dependent rather 
than universal. Even in the remaining configurations where slight degradation 
occurs, the average drop is only 1.8\% absolute while the efficiency gains. 
Section 4.3 provides empirical grounding: probing analysis confirms that low-impact-score layers (high substitutability) 
are replaced early and high-impact-score layers preserved, validating the algorithm's implicit prioritization.

\subsubsection{Performance-Efficiency Trade-offs.}
\begin{figure}[!h]
\centering
\includegraphics[width=\textwidth]{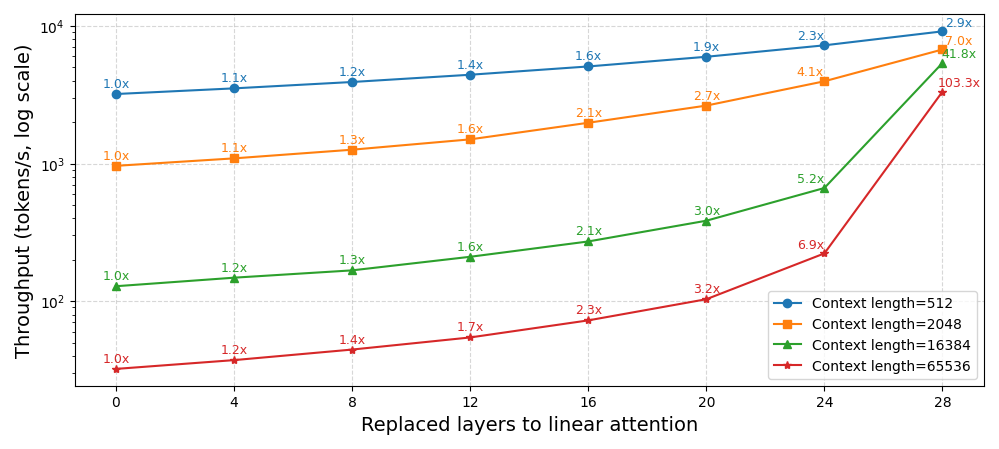}
\caption{Throughput comparison under context length of 512, 2,048, 16,384 and 65,536. The numbers above points indicate the speedup relative to base full-attention model. } \label{fig2}
\end{figure}

As previously noted, incorporating more linear attention modules can lead to higher inference speed; however, this often comes at the cost of reduced accuracy. 
Using Llama3.2-3B-Instruct as the base model, we compare decoding throughput across configurations with different numbers of layers replaced by Jet-Block, under context lengths of 512, 2,048, 16,384 and 65,536, following same evaluation protocol in \cite{gu2025jet}. 
As illustrated in Fig.~\ref{fig2}, decoding throughput increases with the number of linear layers, and greater speedups are observed at longer context lengths.

\begin{table}[!htbp]
\centering
\caption{Performance of base model ($\mathcal{P}_{\mathrm{base}}$) and optimized model ($\mathcal{P}_{\mathrm{opt}}$), showing
performance drop percent ($Drop$ (\%)), number of replaced layers ($\#R$) and corresponding throughput speedup($\times$) under context lengths of 2,048.}
\label{tab2}
\begin{adjustbox}{width=\textwidth}
\begin{tabular}{lcccccccccccccccccc}
\toprule
 & ArcC & ArcE & BQ & CQ & HS & OBA & PIQA & WG & GSM8K & MQ & HE & MBPP & SWDE & SAMSum & TREC & EC & MK & PM \\
\hline
$\mathcal{P}_{\mathrm{base}}$ & 46.4 & 67.9 & 78.6 & 67.7 & 52.4 & 36.0 & 75.5 & 68.0 & 63.9 & 34.5 & 51.2 & 54.8 & 85.8 & 42.4 & 72.0 & 39.5 & 86.8 & 69.8 \\
$\mathcal{P}_{\mathrm{opt}}$ & 43.3 & 66.5 & 76.5 & 64.7 & 49.7 & 34.8 & 71.9 & 64.1 & 59.6 & 33.5 & 48.8 & 52.7 & 81.8 & 40.0 & 68.5 & 35.7 & 79.4 & 68.8 \\
$Drop$ & 6.68 & 2.06 & 2.67 & 4.43 & 5.15 & 3.33 & 4.77 & 5.74 & 6.73 & 2.90 & 4.69 & 3.83 & 4.66 & 5.66 & 4.86 & 9.62 & 8.53 & 1.43 \\
$\#R$ & 10 & 11 & 16 & 11 & 9 & 13 & 14 & 12 & 6 & 10 & 6 & 5 & 11 & 10 & 10 & 10 & 10 & 15 \\
$Speedup$($\times$) & 1.47 & 1.54 & 2.13 & 1.54 & 1.38 & 1.42 & 1.86 & 1.61 & 1.24 & 1.47 & 1.24 & 1.17 & 1.54 & 1.47 & 1.47 & 1.47 & 1.47 & 1.98 \\
\bottomrule
\end{tabular}
\end{adjustbox}
\end{table}

We allow a maximum performance drop of 5\% relative to the base model on the validation set and select the hybrid model with the most linear attention layers as the optimal configuration.
Then we examine the resulting trade-off between performance and efficiency on each test dataset.
As shown in Table~\ref{tab2}, $\mathcal{P}_{\text{base}}$ and $\mathcal{P}_{\text{opt}}$ denote the performance of the base and optimal hybrid models, respectively; $Drop$ represents the degradation percent of $\mathcal{P}_{\text{opt}}$ compared with  $\mathcal{P}_{\text{base}}$; and $\#R$ indicates the number of replaced layers to linear attention in the optimal model. 
%The observed performance degradation aligns closely with the tolerated margin. 
The hybrid model achieves substantial speedups with controlled performance degradation (average drop 4.8\%).
With more layers replaced to linear attention,the hybrid model achieves substantially higher inference throughput, a trend that is consistently visualized in Fig.~\ref{fig2}. 
By prescribing an acceptable degradation margin, the performance and efficiency can be flexibly balanced to meet diverse real-world deployment constraints.

\subsubsection{Replacement Order Analysis.}
We investigate the impact of layer replacement trajectories using different base models, linear attention variants, and downstream tasks.

\begin{figure}[!htbp]
\centering
\includegraphics[width=\textwidth]{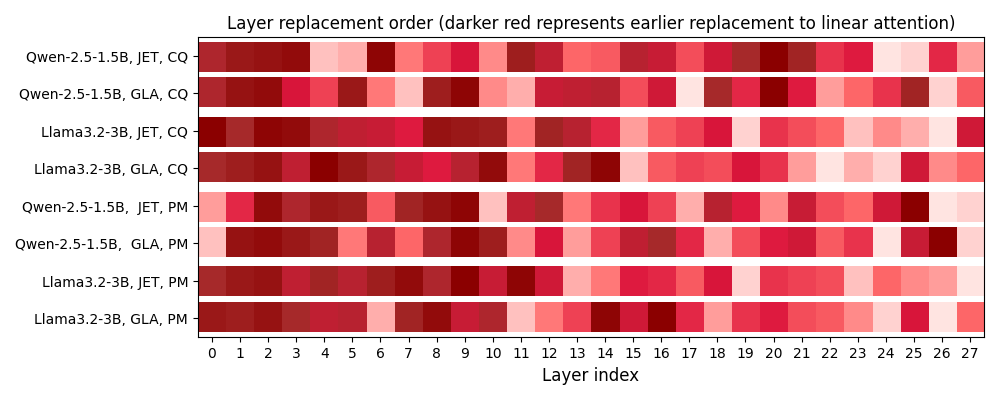}
\caption{Layer replacement trajectories on PubmedQA (PM) and CommonsenseQA (CQ) using Qwen2.5-1.5B and Llama3.2-3B-Instruct (28 layers each)  with linear attention variants: Gated Linear Attention (GLA) and Jet-Block (JET).} \label{fig3}
\end{figure}

As depicted in Fig.~\ref{fig3}, we progressively replace all layers with linear attention by our DtR method.
The result suggests that the parametric storage of knowledge is distributed heterogeneously across different base models, while downstream tasks exhibit distinct knowledge demands.  
Consequently, it is essential to identify a task-specific model architecture.

Interestingly, we observe that the layer replacement trajectories across different linear attention variants exhibit a remarkable degree of consistency.
This suggests that the optimal placement order of linear attention modules is primarily governed by the base model architecture and task characteristics, rather than the specific design of the linear attention mechanism itself.

\begin{figure}[!hbtp]
\centering
\includegraphics[width=\textwidth]{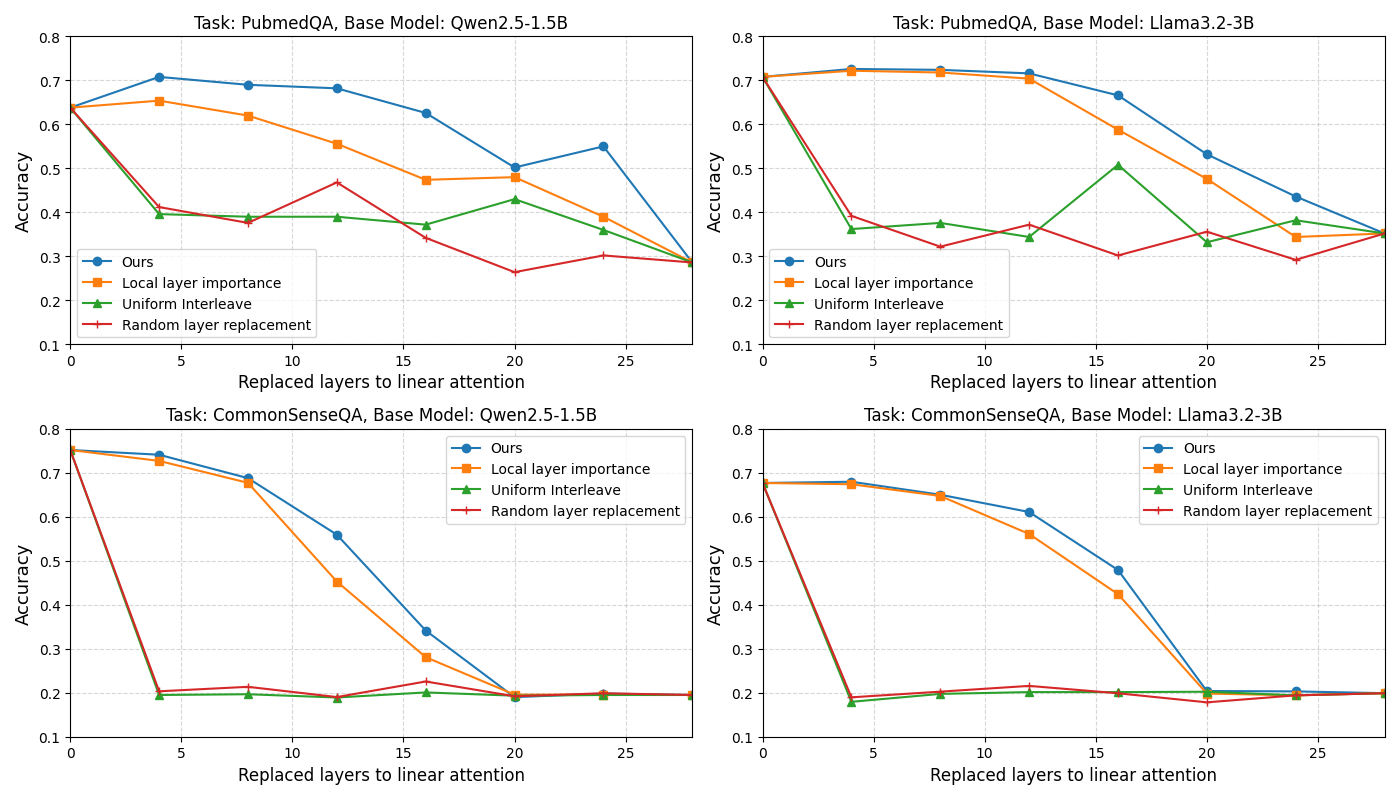}
\caption{Comparison of different replacement strategies.} \label{fig4}
\end{figure}
\vspace{-20pt}

\subsubsection{Comparison with Alternative Replacement Strategies.}
We compare our proposed DtR against several alternative layer replacement strategies:
(1) \textbf{Local layer importance}: layers are replaced one by one according to their importance scores, which are computed by ranking the performance drop caused by replacing each layer to linear attention locally in the base model;
(2) \textbf{Uniform interleaving}: full-attention and linear-attention layers are evenly interleaved across the model depth;
(3) \textbf{Random layer replacement}: in each iteration, layers to be replaced with linear attention are selected at random.

The results are shown in Fig.~\ref{fig4}. Our replacement strategy consistently and significantly outperforms other methods across varying numbers of replaced layers, two distinct base models and tasks, highlighting the effectiveness of greedy, validation-guided replacement in identifying task-specific hybrid architectures.

\subsubsection{Experimental Cost Analysis.}
We report the total experimental cost of our method on a single NVIDIA A800 GPU on the PubMedQA dataset with different base models.
Note that the greedy search ends when all full-attention layers are replaced, providing an upper-bound greedy search runtime.
Table~\ref{tab3} shows that the entire pipeline completes in just a few hours on a single A800 GPU, enabling efficient conversion from a pretrained model into a task-specific hybrid architecture.
The data requirement is also minimal—approximately 100M general-domain tokens for blockwise local distillation and only a small task-specific validation set.

\begin{table}
\centering
\caption{GPU hours across base models during BLD and greedy replacement on PubMedQA, adopting 100M general-domain tokens for blockwise local distillation and 500 samples for task-specific validation.}
\label{tab3}
\begin{adjustbox}{max width=\textwidth, center}
\begin{tabular}{lccc}
\toprule
Base Model & \makecell{BLD\\GPU Hours} & \makecell{Greedy replacement\\GPU Hours} & \makecell{Total\\GPU Hours} \\
\midrule
Qwen2.5-1.5B & 2.5 & 2.5 & 5.0 \\
Llama3.2-3B-Instruct & 4.0 & 3.0 & 7.0 \\
Llama3.1-8B-Instruct & 8.0 & 7.0 & 15.0 \\
\bottomrule
\end{tabular}
\end{adjustbox}
\end{table}
\vspace{-20pt}

\begin{figure}[!htbp]
\centering
\includegraphics[width=\textwidth]{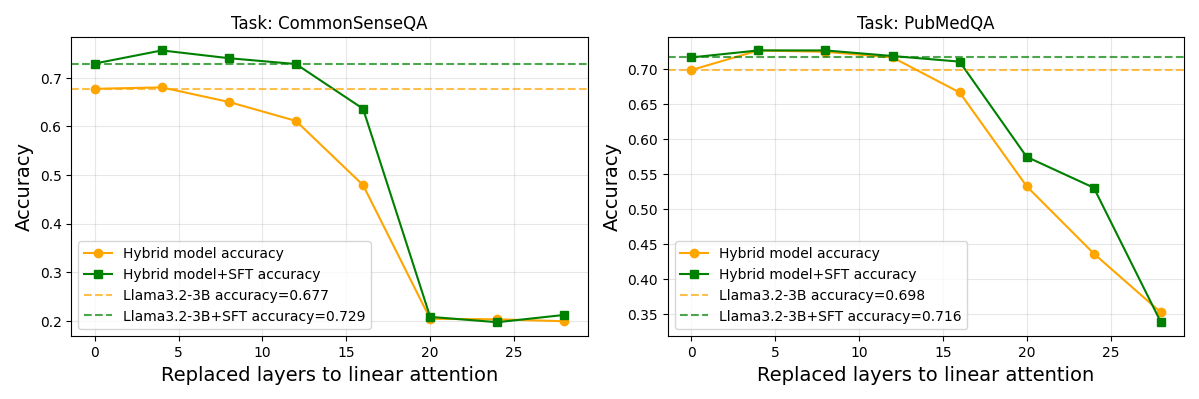}
\caption{Supervised fine-tuning (SFT) on DtR-searched hybrid model for further gains.} \label{fig5}
\end{figure} 
\vspace{-20pt}

\subsubsection{Fine-Tuning for Further Gains on DtR-searched Hybrid Model.}
Building upon the best hybrid models obtained through our DtR procedure, we investigate the impact of supervised fine-tuning (SFT) on downstream task performance, using Llama3.2-3B-Instruct (28 layers) as the base model and Jet-Block as the linear attention module. 
We apply SFT to the official training data of CommonsenseQA and PubMedQA for each best hybrid model, varying the number of replaced attention layers across \{4, 8, 12, 16, 20, 24, 28\}.

As shown in Fig.~\ref{fig5}, our DtR-searched hybrid models either match or surpass the performance of the base model with fewer than 12 to 16 layers replaced by linear attention. Moreover, when supervised fine-tuning is applied to these hybrid configurations, their performance improves further—sometimes even surpassing that of the fine-tuned base model. This demonstrates that our DtR-searched hybrid architectures can effectively benefit from additional improvement mechanisms such as fine-tuning.

\vspace{-10pt}
\subsection{Mutual Information based Empirical Analysis}
Following the Information Bottleneck principle in \cite{alain2016understanding}, we operationalize mutual information estimation without explicit density modeling by treating linear probe accuracy as a practical lower bound on $I(\mathbf{H};Y)$  —the mutual information between hidden states $\mathbf{H}$ and task labels \(Y\). Building on this, we design a probing-based criterion to identify which full-attention layers can be replaced by their linear counterparts without incurring performance loss.  
\vspace{-10pt}
\subsubsection{Impact Score.} For each layer \(\ell\), let \(A_{\text{full}}^{(\ell)}\) be the accuracy of a linear probe trained on the full module's hidden states \(\mathbf{H}_{\text{full}}^{(\ell)}\) to predict task label \(Y\). Similarly, \(A_{\text{linear}}^{(\ell)}\) is the probe accuracy on the linear module's states \(\mathbf{H}_{\text{linear}}^{(\ell)}\), and \(A_{\text{cross}}^{(\ell)}\) is the accuracy of the full-module probe evaluated on \(\mathbf{H}_{\text{linear}}^{(\ell)}\). Here, \(A_{\text{linear}}\) measures task-relevant information retained by the linear module, while \(A_{\text{cross}}\) measures alignment with the full module's discriminative directions.

A good replacement candidate should have both \(A_{\text{linear}}\) and \(A_{\text{cross}}\) close to \(A_{\text{full}}\). Thus we define a layer-wise \textbf{impact score} that is high when replacement is harmful (i.e., the layer should be kept as full attention):

\[
\text{Impact Score}^{(\ell)} = 1 - \frac{A_{\text{linear}}^{(\ell)} + A_{\text{cross}}^{(\ell)}}{2 A_{\text{full}}^{(\ell)}}.
\]

Intuitively, if linear module perfectly preserves both task information and alignment (\(A_{\text{linear}} \approx A_{\text{cross}} \approx A_{\text{full}}\)), then \(\text{impact score} \approx 0\) — the layer should be replaced early. If the linear module performs poorly on either metric, importance becomes positive, indicating the layer should be retained. The score is monotonic with respect to the greedy replacement order: layers with lower importance are replaced earlier.

\begin{figure}[!htbp]
    \centering
    \includegraphics[width=\textwidth]{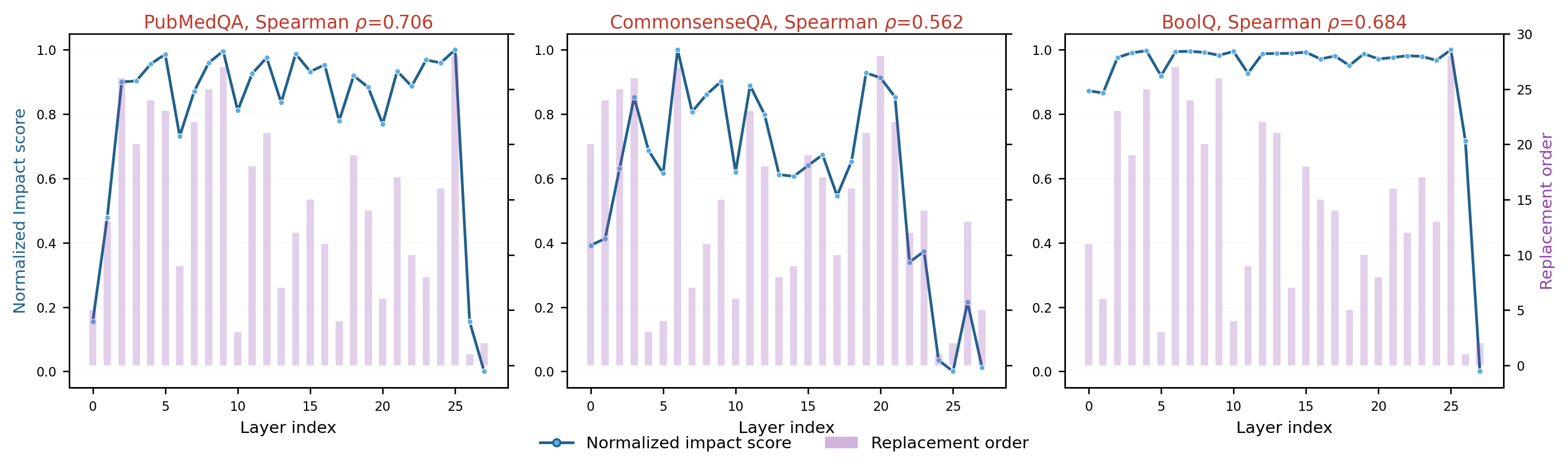}
    \caption{Visualization between the impact score (normalized from 0 to 1) and actual replacement order (taller = replaced later,  should be kept as full attention) across three tasks(PubMedQA, CommonsenseQA and BoolQ), using Qwen2.5-1.5B (28 layers) as the base model and Jet-Block as the linear attention module.}
    \label{fig6}
\end{figure}

We validate this by computing the Spearman rank correlation between the normalized \(\text{Impact Score}^{(\ell)}\) and the actual replacement order (1 for earliest replaced, \(L\) for last). Fig.~\ref{fig6} shows consistently high positive correlations across tasks, confirming that our probing score reliably predicts replacement. This provides theoretical grounding for our method: the algorithm implicitly prioritizes layers with low impact score (high linear substitutability) for replacement.

\vspace{-20pt}

\subsubsection{Fixed-ratio layer replacement analysis: why validation feedback matters.}
While Impact Score explain replacement order post-hoc, whether they can prescribe optimal configurations remains open. As Table~\ref{tab4} shown, the two configurations diverge significantly: Impact-Score-Guided's 10-layer setup underperforms Greedy-Layer-Replacement with the same budget. This reveals \textbf{layer interactions}: replacing one layer alters the functional role of others. 
Validation feedback captures these dependencies, while static impact score cannot. The performance gap quantifies the cost of skipping task-specific validation, justifying greedy search's modest overhead.

\begin{table}[!htbp]
\centering
\small
\caption{Greedy layer replacement outperforms static Impact-Score guidance under fixed replacement budget (10/28 layers, Qwen2.5-1.5B, Jet-Block). }
\label{tab4}
\begin{adjustbox}{max width=\textwidth, center}
\begin{tabular}{lcccc}
\toprule
Task & Method & Replaced Layers & Acc (\%) & $\Delta$ \\
\midrule
\multirow{2}{*}{PubMedQA} 
 & Greedy Layer Replacement & \{4,8,12,19,21,22,23,24,25,26\} & \textbf{70.2} & -- \\
 & Impact-Score-Guided & \{0,6,10,13,17,20,22,23,26,27\} & 66.4 & -3.8 \\
\midrule
\multirow{2}{*}{CommonsenseQA} 
 & Greedy Layer Replacement & \{2,3,6,10,14,19,23,24,25,26\} & \textbf{64.8} & -- \\
 & Impact-Score-Guided & \{4,5,7,10,13,14,17,24,25,27\} & 61.3 & -3.5 \\
\midrule
\multirow{2}{*}{BoolQ} 
 & Greedy Layer Replacement & \{2,5,10,18,19,22,23,24,25,26\} & \textbf{73.0} & -- \\
 & Impact-Score-Guided & \{1,5,10,11,14,18,19,20,26,27\} & 70.4 & -2.6 \\
\bottomrule
\end{tabular}
\end{adjustbox}
\end{table}
\vspace{-30pt}

\section{Conclusion}  
We have presented a practical and efficient framework for automatically constructing task-specific hybrid models that judiciously integrates full attention and  linear attention mechanisms. 
By first distilling linear attention counterpart for each full attention block via blockwise local distillation, and then applying a greedy, validation-guided replacement strategy, our proposed DtR identifies a high-performing hybrid architecture without costly pretraining or neural architecture search.
Extensive experiments show that our approach yields task-specific hybrid models, which strategically retain full attention in task-critical layers to preserve expressive power, while adopting linear attention elsewhere to gain computational efficiency.

Crucially, our approach is both backbone-agnostic and task-general: it can be seamlessly applied to any pretrained full-attention models and adapted to diverse downstream tasks with minimal overhead. 
We believe this paradigm offers a scalable and deployment-friendly pathway toward efficient adaptation of foundation models, particularly in resource-constrained or latency-sensitive scenarios.

\tiny
\bibliographystyle{splncs04}
\bibliography{ref}
\end{document}